\documentclass[10pt,journal,compsoc]{IEEEtran}

\usepackage{multirow}
\usepackage[table]{xcolor}
\usepackage{algorithm} 
\usepackage{amssymb}
\usepackage{amsfonts}
\usepackage{multicol}
\usepackage{graphicx}
\usepackage{booktabs}
\usepackage[T1]{fontenc}
\usepackage[breaklinks=true,bookmarks=true,pagebackref=false,colorlinks,linkcolor=blue,anchorcolor=red,citecolor=blue,urlcolor=blue]{hyperref}
\usepackage[numbers,sort]{natbib}

\usepackage{comment}
\usepackage{caption}
\usepackage{subcaption}
\usepackage{amsmath}
\usepackage{algorithmic}
\usepackage{array}
\usepackage[switch]{lineno}
\usepackage{longtable}
\usepackage{xspace}
\usepackage{url}
\usepackage{overpic}
\usepackage{ragged2e}
\usepackage{xpatch}
\usepackage{pifont}
\usepackage{enumitem}
\usepackage{bbm}
\usepackage{colortbl}

\usepackage[capitalize]{cleveref}
\crefname{section}{Sec.}{Secs.}
\Crefname{section}{Section}{Sections}
\Crefname{table}{Table}{Tables}
\crefname{table}{Tab.}{Tabs.}

\newcommand{\algname}{TopoNet\xspace}
\newcommand{\datasetname}{OpenLane-V2\xspace}

\definecolor{hanblue}{RGB}{68, 114, 196}
\definecolor{cangopink}{RGB}{255, 126, 121}
\definecolor{blood}{RGB}{148, 17, 0}

\makeatletter
\DeclareRobustCommand\onedot{\futurelet\@let@token\@onedot}
\def\@onedot{\ifx\@let@token.\else.\null\fi\xspace}
\def\eg{e.g\onedot} 
\def\ie{i.e\onedot}

\def\etal{et~al\onedot}

\makeatother

\definecolor{darkgreen}{rgb}{0,0.7,0}
\definecolor{darkblue}{RGB}{31,119,180}
\definecolor{darkred}{RGB}{214,39,40}
\definecolor{mediumgray}{rgb}{0.5,0.5,0.5}
\definecolor{mediumteal}{rgb}{0,0.5,0.5}

\definecolor{ellisred}{rgb}{0.87,0.44,0.38} 
\definecolor{ellisgreen}{rgb}{0.69,0.90,0.52} 
\definecolor{elliscyan}{rgb}{0.29,0.77,0.74} 
\definecolor{ellisorange}{rgb}{0.89,0.55,0.28} 
\definecolor{ellisblue}{rgb}{0.41,0.61,0.86} 

\begin{document}
\title{
Graph-based Topology Reasoning \\for Driving Scenes
}

\author{
    Tianyu Li$^{*}$,
    Li Chen$^{*\dagger}$,
    Huijie Wang,
    Yang Li, 
    Jiazhi Yang,
    Xiangwei Geng,
    Shengyin Jiang, \\
    Yuting Wang,
    Hang Xu,
    Chunjing Xu,
    Junchi Yan,
    Ping Luo
    and Hongyang Li$^{\dagger}$

\IEEEcompsocitemizethanks{
\IEEEcompsocthanksitem $*$ equal contribution. $\dagger$ project lead.
\IEEEcompsocthanksitem
Primary contact: \texttt{hy@opendrivelab.com}
\IEEEcompsocthanksitem T. Li, L. Chen, H. Wang, Y. Li, J. Yang and H. Li are with OpenDriveLab, Shanghai AI Lab, China. 
H. Xu and X. Xu are with Huawei. 
T. Li is also affiliated with Fudan University. 
L. Chen and P. Luo are with The University of Hong Kong. 
J. Yan and H. Li are with Shanghai Jiao Tong University, China.
}
\thanks{Manuscript received August, 2023.}}

\IEEEtitleabstractindextext{%
\begin{abstract}
\justifying 
   Understanding the road genome is essential to realize autonomous driving.
   This highly intelligent problem contains two aspects - the connection relationship of lanes, and the assignment relationship between lanes and traffic elements, where a comprehensive topology reasoning method is vacant.
   On one hand, previous map learning techniques struggle in deriving lane connectivity with segmentation or laneline paradigms; or prior lane topology-oriented approaches focus on centerline detection and neglect the interaction modeling. 
   On the other hand, the traffic element to lane assignment problem is limited in the image domain, leaving how to construct the correspondence from two views an unexplored challenge.
   To address these issues, we present \textbf{\algname}, the first end-to-end framework capable of abstracting traffic knowledge beyond conventional perception tasks.
   To capture the driving scene topology, we introduce three key designs: (1) an embedding module to incorporate semantic knowledge from 2D elements into a unified feature space; (2) a curated scene graph neural network to model relationships and enable feature interaction inside the network; (3) instead of transmitting messages arbitrarily, a scene knowledge graph is devised to differentiate prior knowledge from various types of the road genome.
   We evaluate \algname on the challenging scene understanding benchmark, \datasetname, where our approach outperforms all previous works by a great margin on all perceptual and topological metrics.
   The code is released at \url{https://github.com/OpenDriveLab/TopoNet}
\end{abstract}

}

\maketitle

\IEEEdisplaynontitleabstractindextext

%
\IEEEpeerreviewmaketitle

\IEEEraisesectionheading{
\section{Introduction}\label{sec:introduction}
}

\begin{figure}[t!]
  \centering
  \includegraphics[width=.95\linewidth]{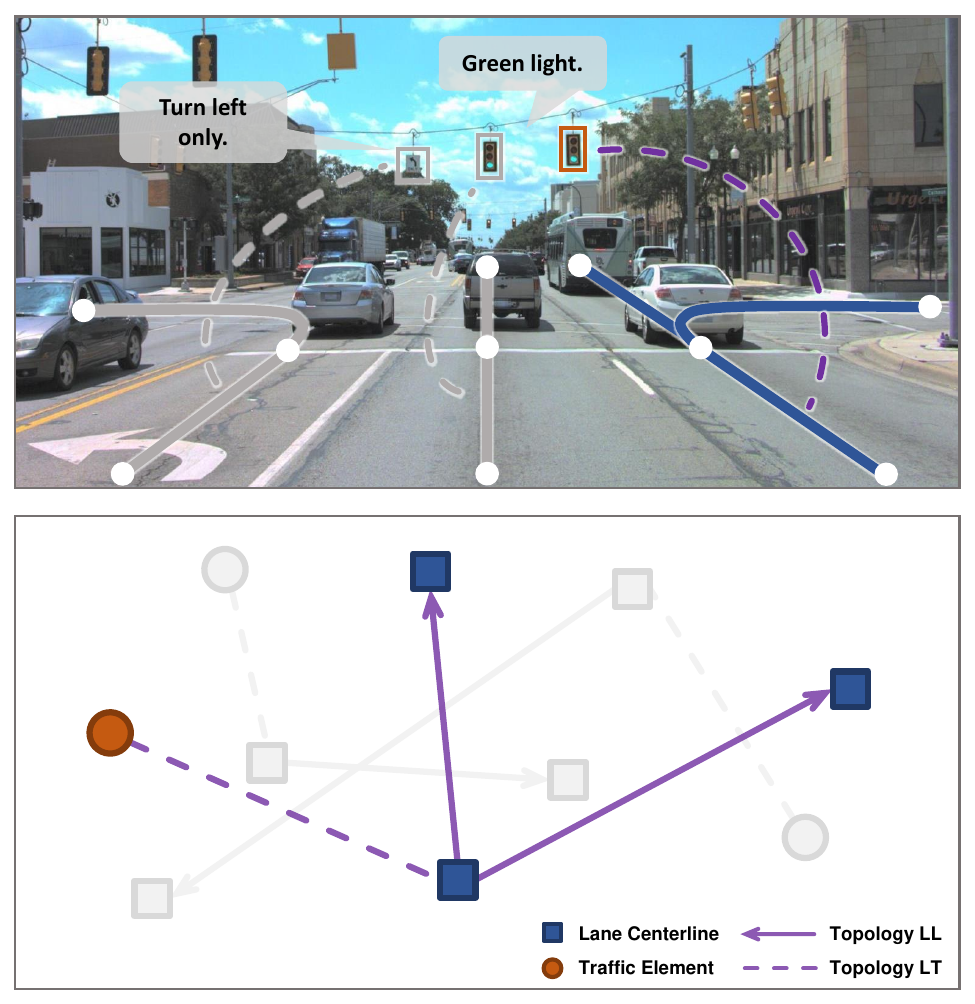}
  \caption{
    \textbf{Topology relationship of driving scenes.}
    While driving into an intersection, the self-driving vehicle has to reason about the correct lane and traffic information for downstream navigation. 
    We advocate, and present \algname, to directly achieve  topology understanding on the heterogeneous graph.
    ``Topology LL'' and ``Topology LT'' represent the relationship among lane centerlines and the relationship between lane centerlines and traffic elements respectively.
  }
  \label{fig:motivation}
\end{figure}

\IEEEPARstart{I}{magine} that an autonomous vehicle is navigating towards a complex intersection and planning to go straight: it is wondering which one of the lanes in front to drive into and which traffic signal to follow. This high-level intellectual problem requires the agent not only to perceive lane position accurately, but also to understand the topology relationship from sensor inputs.
Specifically, the driving scene topology includes: (1) the \textbf{lane topology graph} comprising centerlines as well as their connectivity, (2) and the \textbf{assignment relationship} between lanes and traffic elements (\eg, traffic lights, traffic boards, and road markers). 
As illustrated in \cref{fig:motivation}, they altogether build a topological structure that provides explicit navigation signals for downstream tasks such as motion prediction and planning~\cite{bansal2018chauffeurnet, chai2020multipath}.

Conventional autonomous driving datasets~\cite{caesar2020nuscenes, wilson2av2} include lane topology implicitly in the High-Definition (HD) map, which is designed for mapping but not being learned by neural networks.
Various formulations are proposed to serve as substitutes to HD maps, such as 2D and 3D laneline detection~\cite{pan2018culane,tabelini2021laneatt,chen2022persformer, 3dlanenet, guo2020genlanenet}, bird's-eye-view (BEV) segmentation~\cite{pan2020vpn,roddick2020predicting,li2021hdmapnet, xu2022centerlinedet}, and BEV map vecterization~\cite{liu2022vectormapnet,liao2023maptr, liao2023maptrv2}.
To derive lane connecticity, a ``tabula-rasa'' resolution is to directly average two neighboring lanelines to get centerlines and then connect them into a lane graph, based on the instance-wise lanelines in vectorization-based methods.
Yet, it demands complicated hand-crafted rules and heavy burdens on post-processing.
Another approach is to supervise the original perception frameworks with relationship labels. 
Recent studies, STSU~\cite{can2021stsu} and TPLR~\cite{can2022tplr}, employ a Transformer-based architecture to predict lane instances and an additional MLP to output their connectivity. 
But still, they suffer from overfitting to surrounding background characteristics or difficulty in finding useful information without explicit relationship modeling inside the network.

Moreover, the relationship assignment problem between traffic elements and lanes from sensor inputs remains mostly unexplored.
Previous work~\cite{langenberg2019deepmetafusion} tried to associate the ground truth of lanelines and traffic lights in the image domain (perspective view, PV).
However, integrating traffic elements and lanes in a heterogenous graph (\cref{fig:motivation}) is a different story.
A reason is that traffic elements are described as bounding boxes in PV, while lanes are characterized as curves in 3D or BEV space.
Meanwhile, spatial locations remain less important for traffic elements as their semantic meanings are essential, but positional clues of lanes are crucial for autonomous driving vehicles.

To address these issues, we present a Topology Reasoning Network (\textbf{\algname}), which predicts the driving scene topology in an end-to-end manner. 
As an attempt to reason about scene topology in a single network, \algname comprises two branches with a shared feature extractor, for traffic elements and centerlines respectively.
Motivated by the Transformer-based detection algorithms~\cite{carion2020detr, zhu2020deformabledetr}, we employ instance queries to extract local features via the deformable attention mechanism, which restricts the attention region and accelerates convergence.
Since the clues for locating a specific centerline instance could be encoded in its neighbors and corresponding traffic elements, a Scene Graph Neural Network (SGNN) is devised to transmit messages among instance-level embeddings.
Furthermore, we propose a scene knowledge graph to capture prior topological knowledge from entities of different types.
Specifically, a series of GNNs are developed based on categories of traffic elements and the centerline connectivity relationship (\ie, predecessor, ego, successor). Updated queries are ultimately decoded as the perception results and driving scene topology.
With the proposed designs, we substantiate \algname on the large-scale topology reasoning benchmark, \datasetname~\cite{openlanev2}. \algname outperforms state-of-the-art approaches by 15-84\% for centerline perception, and achieves times of performance in terms of the challenging topology reasoning task. Ablations are conducted to verify the effectiveness of our framework.

To sum up, our key contributions are as follows:
\begin{itemize}
    \item We present \algname, which unifies and enhances heterogeneous feature learning via graph-based design. The superiority of \algname is demonstrated on the challenging topology reasoning benchmark with a wide margin on perceptual and topological metrics.
    \item Through the elaborated graph-based modules, the position and shape of lanes are refined by neighboring centerlines and assigned traffic elements. We conduct exhaustive experiments to validate the message-passing mechanism.
    \item Drawing inspiration from the proposed network and our philosophy, a collection of innovative methods emerged in the CVPR 2023 Autonomous Driving Challenge\footnote{\url{https://opendrivelab.com/AD23Challenge.html}}. The \algname and works from participants have made joint contributions to shape the community.
\end{itemize}

\section{Related Work}
\label{sec:related}

\subsection{Lane Graph Learning} 
Lane Graph Learning has received abundant attention due to its pivotal role in autonomous driving. 
Prior works investigate generating road graphs~\cite{he2020sat2graph, bandara2022spin} or spatially denser lane graphs~\cite{homayounfar2019dagmapper, zurn2021lane, he2022lane, buchner2023lanegnn} from aerial images. However, roads in aerial images are often occluded by trees and buildings, leading to inaccurate results.
Recently, there has been a growing focus on producing lane graphs directly from vehicle-mounted sensor data. 
STSU~\cite{can2021stsu} proposes a DETR-like neural network to detect centerlines
and then derive their connectivity by a successive MLP module. 
Based on STSU, Can~\etal~\cite{can2022tplr} introduce additional minimal cycle queries to ensure proper order of overlapping lines.
CenterLineDet~\cite{xu2022centerlinedet} regards centerlines as vertices and designs a graph-updating model trained by imitation learning. 
It is also worth noticing that Tesla proposes the ``language of lanes'' to represent the lane graph as a sentence~\cite{tesla2022aiday}. 
The attention-based model recursively predicts lane tokens and their connectivity.
In this work, we focus on explicitly modeling the centerline connectivity inside the network to enhance feature learning and indulging traffic elements in constructing the full driving scene graph.

\subsection{HD Map Perception}
With the trending popularity of BEV perception~\cite{philion2020lss, li2022bevformer, zhou2022cross, uniad, gao2023sdf, liao2023maptrv2}, recent works focus on learning HD Maps with segmentation and vectorized methods.
Map segmentation aims at predicting the semantic meaning of each BEV grid, such as lanelines, pedestrian crossings, and drivable areas. These works differentiate from each other mainly in the perspective view to BEV transform module, \ie, IPM-based~\cite{xie2022m2bev, can2022understanding}, depth-based~\cite{liu2022bevfusion, hu2022stp3}, or Transformer-based~\cite{li2022bevformer, jiang2022polarformer}. 
Though dense segmentation provides pixel-level information, it cannot touch down the complex relationship of overlapping elements. Li~\etal~\cite{li2021hdmapnet} handles the problem by grouping and vectorizing the segmented map with complicated post-processings. 
VectorMapNet~\cite{liu2022vectormapnet} proposes to directly represent each map element as a sequence of points, which uses coarse key points to decode laneline locations sequentially.
MapTR~\cite{liao2023maptr} further explores a unified permutation-based modeling approach for the sequence of points to eliminate the modeling ambiguity and improve performance and efficiency.
In fact, since vectorization also enriches the direction information for lanelines, vectorization-based methods could be easily adapted to centerline perception by alternating the supervision. 
Recently, InstaGraM~\cite{shin2023instagram} constructs map elements as a graph by predicting vertices first and then utilizing a GNN module to detect edges. Its GNN produces all vertex features simultaneously, leading to the lack of instance-level interaction.
Contrary to the aforementioned approaches, we leverage instance-wise feature transmission with a graph neural network, to extract prominent prediction hints from other elements in the topology graph.

\subsection{Driving Scene Understanding} 
Driving Scene Understanding mainly indicates summarizing positional relationships of elements in outdoor environments beyond perception~\cite{zipfl2022towards, tian2020road, mylavarapu2020understanding, malawade2022roadscene2vec}. 
Previous works focus on utilizing the relationships of 2D bounding boxes for motion prediction~\cite{li2020learning, mylavarapu2020towards, mylavarapu2020understanding, fang2022heterogeneous} and risk assessment~\cite{yu2021scene, malawade2022spatiotemporal}. 
In the industrial context, Mobileye presents an optimization-based method to automatically construct lane topology and traffic light-to-lane relationships based on their internal data~\cite{mobileye2022ces}. 
In the academy, Langenberg~\etal~\cite{langenberg2019deepmetafusion} address the traffic light to lane assignment~(TL2LA) problem with a convolutional network by taking heterogeneous metadata as additional inputs.
In contrast, \algname takes RGB images only and additionally reasons about the topology for lane entities besides TL2LA. We instantiate \algname on the large-scale driving scene understanding benchmark, which covers complicated urban scenarios.

\subsection{Graph Neural Network} 
Graph Neural Network and its variants, such as graph convolutional network (GCN)~\cite{kipf2016semi}, GraphSAGE~\cite{hamilton2017inductive}, and GAT~\cite{velivckovic2017graph}, are widely adopted to aggregate features of vertices and extract information from graph data~\cite{scarselli2008graph}.
Witnessing the impressive achievements of GNN in various fields (\eg, recommendation system and video understanding)~\cite{guo2020deep, pradhyumna2021graph, chang2021comprehensive, mohamed2020social}, researchers in the autonomous driving community also attempt to utilize it to process unstructured data. 
Weng \etal~\cite{weng2020gnn3dmot, weng2021ptp} introduce GNN to capture interactions among agent features for 3D multi-object tracking.
LaneGCN~\cite{liang2020lanegcn} constructs a lane graph from HD map, while others~\cite{fang2022heterogeneous, jia2022hdgt, jia2022temporal} model the relationship of moving agents and lanelines as a graph to improve the trajectory forecasting performance.
Inspired by prior works, we design a GNN for the driving scene understanding task to enhance feature interaction and introduce a class-specific knowledge graph to better incorporate semantic information.

\begin{figure*}[t]
  \centering
  \includegraphics[width=.9\linewidth]{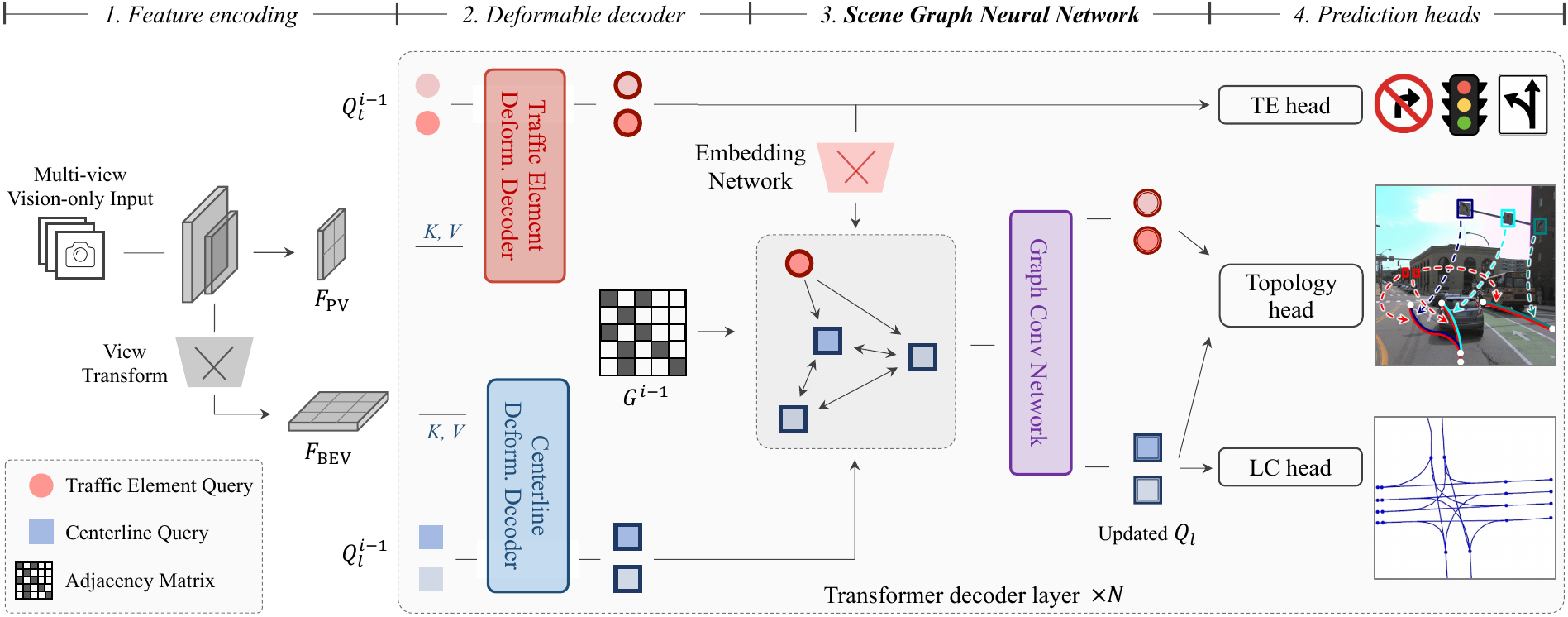}
  \caption{
  \textbf{Systematic diagram of \algname}. \algname addresses the crucial problem of topology reasoning for driving scenes in an end-to-end fashion. It consists of four stages, with the last three being compacted in a Transformer decoder architecture.
  \algname handles traffic elements and centerlines as two parallel branches at the Deformable decoder stage.
  Various types of instance queries (\textcolor{cangopink}{red}, \textcolor{hanblue}{blue}) then interact, exchange messages, acquire and aggregate prominent knowledge in the proposed Scene Graph Neural Network stage.
  The explicit relationship modeling inside the network serves as a favorable scheme for feature learning and topology prediction.
    We abbreviate traffic elements and lane centerlines as ``TE'' and ``LC'' in this paper, respectively.
  }
  \label{fig:pipeline}
\end{figure*}

\section{TopoNet}
\label{sec:method}

\subsection{Problem Formulation}

Given images from multi-view cameras mounted on a vehicle, the goal of \algname lies in two perspectives - perceiving entities and reasoning their relationships.
As an instance-level representation is preferable for topology reasoning, a directed centerline is described as an ordered list of points. We denote it as $v_l = [p_0, ..., p_{n-1}]$, where $p = (x, y, z) \in \mathbb{R}^3$ describes  a point's coordinate in 3D space, $p_0$ and $p_{n-1}$ are the starting and ending point respectively.
Traffic elements are represented as 2D bounding boxes in different classes on the front-view images. All existing lanes $V_l$ and traffic elements $V_t$ within a predefined range are required to be detected.

On the perceived entities, the topology relationships are built.
The connectivity of directed lanes establishes a map-like network on which vehicles can drive and is denoted as the lane graph $(V_l, E_{ll})$, where the edge set $E_{ll} \subseteq V_l \times V_l$ is asymmetric. An entry $(i, j)$ in $E_{ll}$ is positive if and only if the ending point of the lane $v_i$ is connected to the starting point of $v_j$.
The graph $(V_l \cup V_t, E_{lt})$ describes the correspondence between lanes and traffic elements. It can be seen as a bipartite graph that positive edges only exist between $V_l$ and $V_t$.
Both edge sets are required to be predicted in the task of topology reasoning.

\subsection{Overview}

\cref{fig:pipeline} illustrates the overall architecture of the proposed \textbf{\algname}.
Given multi-view images as input, the feature extractor generates multi-scale image features, including the front-view feature $F_{\text{PV}}$, and then convert them into a BEV feature $F_{\text{BEV}}$ through a view transform module.
Two independent decoders with the same deformable attention architecture~\cite{zhu2020deformabledetr} consume $F_{\text{PV}}$ and $F_{\text{BEV}}$ to produce instance-level embeddings $Q_t$ and $Q_l$ separately.
The proposed \textbf{Scene Graph Neural Network (SGNN)} then refines centerline queries $Q_l$ in positional and topological aspects.
Note that the decoder and SGNN layers are stacked iteratively to obtain local and global features in a sequential fashion. Finally, the task-specific heads take the refined queries to produce prediction results. Next, we elaborate on the proposed SGNN module.

\subsection{Scene Graph Neural Network}
\label{sec:method-sgnn}

A representative embedding (or query) could provide ideal instance-wise detection or segmentation results, as discussed in conventional perception works~\cite{carion2020detr, wu2022seqformer}.
Nevertheless, being discriminative is not enough to recognize correct topology relationships.
The reason is that it takes a pair of instance queries as input to determine their relationship, in which feature embeddings are actually not independent.
Meanwhile, adopting the local feature aggregation scheme of point-wise queries~\cite{liu2022vectormapnet, liao2023maptr} for centerline perception is inadequate.
Specifically, a key difference between centerlines and physical map elements is that centerlines naturally encode lane topology and traffic rules, which cannot be inferred from local features alone.
Therefore, we aim to simultaneously acquire perception and reasoning results by modeling not only discriminative instance-level representations but also inter-entity relationships.

To this end, we present SGNN, which has several designs and merits compared to previous works. (1) It adopts an embedding network to extract TE knowledge within a unified feature space. (2) It models all entities in a frame as vertices in a graph, and strengthens interconnection among perceived instances to learn their inherent relationships with a graph neural network. (3) Alongside the graph structure, SGNN incorporates prior topology knowledge with a scene knowledge graph. We detailedly introduce each part below.

\subsubsection{Embedding Network}
As traffic elements are labeled on the perspective view, it is hard to harness their positional features in the BEV space.
However, their semantic meaning imposes a great effect that could not be neglected.
For instance, a road sign indicating the prohibition of left turn usually corresponds to lanes that lay in the middle of the road.
This predefined knowledge is beneficial for locating corresponding lanes. Towards this, we introduce an embedding network to extract semantic information and transform it into a unified feature space to match with centerlines:
\begin{equation}
    \widetilde{Q}_t^i = \texttt{embedding}^i\big(Q_t^i\big),
\end{equation}
where $i$ denotes the $i$-th decoder layer. Note that the queries $\widetilde{Q}_t^i$ remain intact in the SGNN.
This is intended since imagining traffic elements from centerlines is relatively challenging, and we empirically find that updating $Q_t$ with too many feature interactions places negative effects on predicting their attributes.

\subsubsection{Feature Propagation in GNN} 
In this part, we introduce how topological relationships are modeled and how knowledge from different queries is exchanged.
Regarding the concept of relationship, GNN is a natural choice. Relations can be conveniently formulated as edges in a graph where entities are seen as vertices, while it is nontrivial in an open world without any explicit constraint.
As there is no prior knowledge of topology structure, a trivial way is to construct a fully connected graph $(V, E)$, where $V = V_l \cup V_t$ and $E \subseteq V \times V$.
This inevitably increases computational cost and introduces unnecessary information transmission, such as between two traffic elements that are placed subjectively by humans.
Instead, we form two directed graphs to propagate features, namely $G_{ll} = (V_l, V_l \times V_l)$ for lane graph estimation and $G_{lt} = (V_l \cup V_t, V_l \times V_t)$ representing the TE to LC assignments.

In graph $G_{ll}$ and $G_{lt}$, lane queries $Q_{l}$ are refined by the connected neighbors and corresponding traffic elements.
Due to the fact that $Q_{l}$ and $Q_{t}$ represent different objects, the semantic gap still exists. We introduce an adapter layer to combine this heterogeneous information into the information gain denoted as $R$.
The overall process in an SGNN layer can be formulated as follows:
\begin{equation}
\begin{aligned}
    Q_{l}^{i'} \ &= \ \texttt{SGNN}_{ll}^i \big( Q_{l}^{i}, G_{ll}^{i-1} \big), \\
    Q_{l}^{i''} &= \ \texttt{SGNN}_{lt}^i \big(Q_{l}^{i}, \widetilde{Q}_{t}^{i}, G_{lt}^{i-1}\big), \\
    R^i \ \  &= \ \texttt{downsample}^{i} \Big(\texttt{ReLU}\big( \texttt{concat}(Q_{l}^{i'}, Q_{l}^{i''}) \big) \Big), \\
    \widetilde{Q}_{l}^{i} \ \ &= \ Q_{l}^{i} + R^i.
\end{aligned}
\end{equation}
Next, we detail how the topological weights in SGNN are constructed and learned, based on a naive scene graph structure and an elaborately designed knowledge graph.

\subsubsection{Vanilla Scene Graph}
Given the adjacency matrix $A_{ll}^{i-1}$, which is a representation of neighboring relationships in graph $G_{ll}^{i-1}$ from the previous layer, we construct a weight matrix $T_{ll}^{i}$ to control the flow of messages in the graph. In the directed graph, messages are passed in a single direction, \eg, from a centerline to its successor.
However, as the structure of lanes depends on each other, the position of a lane is a good indication of the locations of its neighbors.
Thus, we supplement $A_{ll}^{i-1}$ with a backward adjacency matrix to allow message exchange for two connected centerlines.
The matrix $T_{ll}^{i}$ of the $i$-th layer is calculated by:
\begin{equation}
\begin{aligned}
    T^0_{ll} &= I, \\
    T^{i}_{ll} &= \beta_{ll} \cdot \big(A_{ll}^{i-1} + \texttt{transpose}(A_{ll}^{i-1})\big) + I,
\end{aligned}
\end{equation}
where $I$ denotes the identical mapping for self-loop, $\beta_{ll}$ is a hyperparameter to control the ratio of features propagated between nodes. 

In the bipartite graph $G_{lt}$, where only the correspondence between lanes and traffic elements is presented, we utilize features of traffic elements to refine centerline embeddings as follows:
\begin{equation}
\begin{aligned}
    T^0_{lt} &= O, \\
    T^{i}_{lt} &= \beta_{lt} \cdot A_{lt}^{i-1},
\end{aligned}
\end{equation}
where $O$ is a matrix in which all entries are zero.

After obtaining the weight matrices, SGNN utilizes the graph convolutional layer (GCN)~\cite{kipf2016semi} to perform feature propagation among queries:
\begin{equation}
\begin{aligned}
    Q_{l}^{i'} \ &= \texttt{GCN}_{ll}^i \big( Q_{l}^{i}, T_{ll}^{i} \big), \\
    Q_{l}^{i''} &= \texttt{GCN}_{lt}^i \big( Q_{l}^{i}, \widetilde{Q}_{t}^{i}, T_{lt}^{i} \big).
\end{aligned}
\end{equation}

\begin{figure}[t!]
  \centering
  \includegraphics[width=\linewidth]{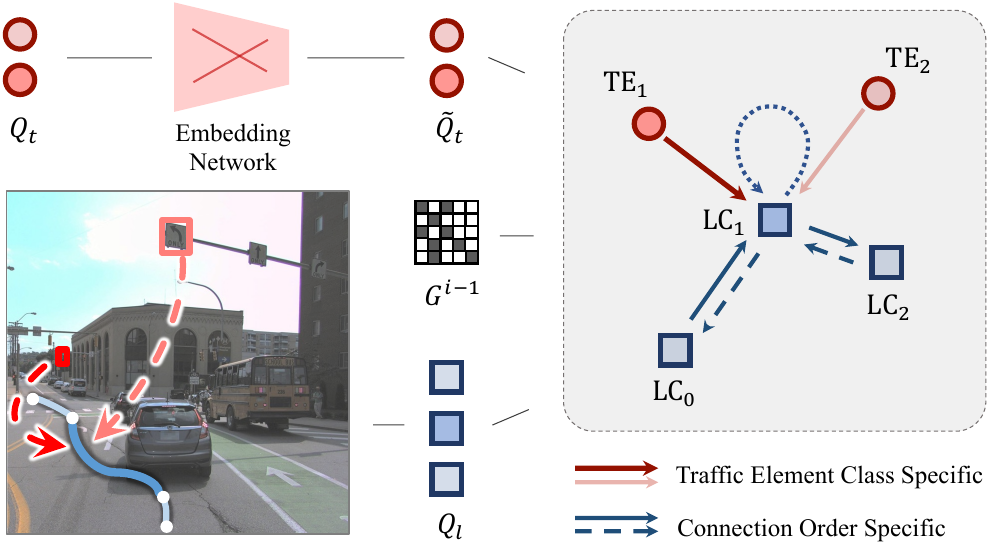}
  \caption{\textbf{Scene knowledge graph} illustration. For the centerline colored \textbf{\textcolor{hanblue}{blue}} in the left case, related weight matrices in the graph are categorically independent. Different traffic elements and lane-directed connections bring different information to the centerline, which is encoded as a scene knowledge graph on the right.
  }
  \label{fig:skg}
\end{figure}

\subsubsection{Scene Knowledge Graph}
Though GCN enables feature propagation in the built graphs and treats nodes differently based on their connectivity, the semantic meaning of vertices remains unexplored.
For example, the information from a traffic element indicating to go straight is not equally important to a red light.
To address the issue and incorporate categorical prior, we design the scene knowledge graph to treat vertices in different classes differently.
\cref{fig:skg} illustrates an example process of updating a centerline query LC$_{1}$ on the given knowledge graph. 

On the graph $G_{lt}$, we use $\mathbf{W}_{lt}^i \in \mathbb{R}^{|C_t| \times F_l \times F_t}$ to denote the learnable weights, where $C_t$ describes the attribute set of traffic elements, $F_l$ and $F_t$ are the number of feature channel of LC and TE queries respectively. A centerline query with index $x$ aggregates information from its corresponding traffic elements based on their classification scores:
\begin{equation}
\begin{aligned}
    K_{lt}^i \ \ &= A_{lt}^{i-1}, \\
    Q_{l_{(x)}}^{i''} &= \sum_{\forall y \in N(x)} \sum_{\forall c_t\in C_t} \beta_{lt} \cdot S^i_{t_{(c_t, y)}} K_{lt_{(x,y)}}^i \mathbf{W}_{lt_{(c_t)}}^{i} {\widetilde{Q}_{t_{(y)}}^{i}},
\end{aligned}
\end{equation}
where $N(x)$ outputs the indices of all neighbors of the vertex with index $x$, and $S^i_{t} \in \mathbb{R}^{|C_t| \times |Q_t^i|}$ represents the classification scores of traffic element queries.

Although all centerlines fall into the same category, the directed connection nature, namely predecessor and successor, still poses an impact on the process of feature propagation.
To this end, we formulate the learnable weight matrix for the lane graph as $\mathbf{W}_{ll}^i \in \mathbb{R}^{|C_l| \times F_l \times F_l}$, where $C_l = \{\text{successor}, \text{predecessor}, \text{self-loop}\}$. The centerline queries are further updated by:
\begin{equation}
\begin{aligned}
    K_{ll}^i \ \ &= \texttt{stack} \big(A_{ll}^{i-1}, \texttt{transpose}(A_{ll}^{i-1}), I \big), \\
    Q_{l_{(x)}}^{i'} &= \sum_{\forall y \in N(x)} \sum_{\forall c_l\in C_l} \beta_{ll} \cdot K_{ll_{(c_l,x,y)}}^i \mathbf{W}_{ll_{(c_l)}}^{i} {Q_{l_{(y)}}^{i}}.
\end{aligned}
\end{equation}

\subsection{Learning}
\label{sec:method-learning}

We employ multiple losses to train \algname in an end-to-end manner. As depicted in \cref{fig:pipeline}, all heads consume queries to provide perception and reasoning results. Nevertheless, they are not entirely independent, as the topology head requires matching results from perception heads.
Similar to Transfomer-based networks~\cite{carion2020detr, zhu2020deformabledetr}, the supervision is applied on each decoder layer to optimize the query feature iteratively.
The overall loss of the proposed model is:
\begin{equation}
\mathcal{L} = \mathcal{L}_{det_{\text{TE}}} + \mathcal{L}_{det_{\text{LC}}} + \mathcal{L}_{top}.
\end{equation}

\subsubsection{Perception}
Following the head design in DETR~\cite{carion2020detr}, the TE head predicts 2D bounding boxes with classification scores. Note that for predicting traffic elements, we take $Q_t$ instead of $\widetilde{Q}_t$ to preserve their positional information in the perspective view.
The LC head produces 11 ordered 3D points and a confidence score from each centerline query $\tilde{q}_l \in \widetilde{Q}_l$. The ground truth of centerlines is normalized based on the predefined BEV range.
For both heads, the Hungarian algorithm is utilized to generate matchings between ground truth and predictions, with the matching cost the same as the loss function. Then task-specific losses $\mathcal{L}_{det_{\text{TE}}}$ and $\mathcal{L}_{det_{\text{LC}}}$ are applied accordingly.
Specifically, for the TE head, we employ the Focal loss~\cite{lin2017focal} for classification, an L1 regression loss, and an IOU loss for localization. Meanwhile, for centerlines, we use Focal loss and L1 loss as the classification and regression loss, respectively.

\subsubsection{Reasoning}
The topology head reasons pairwise relationships on the given embeddings. Similar to STSU~\cite{can2021stsu}, for a pair of instances, we use two MLP layers to reduce the feature dimension for each instance. 
Then the concatenated feature is sent into another MLP with a sigmoid activation to predict their relationship. Based on the matching results from perception heads, the ground truth of each pair of embeddings is assigned.
Different from the TE head, we adopt embeddings from the SGNN module, \ie, the refined queries $\widetilde{Q}_{l}$ for lanes and the semantic embeddings $\widetilde{Q}_t$ for traffic elements. Due to the sparsity of the graph, there exists a severe imbalance in sample distribution, towards which the Focal loss is deployed in $\mathcal{L}_{top}$.

\section{Experiments}
\label{sec:exp}

In this section, we describe experimental settings in detail.
We validate the effectiveness of our design compared to various SOTA methods.
Visualizations are also provided.

\subsection{Protocols}
\label{sec:exp-protocal}

\subsubsection{Implementation Details}

\textbf{Feature Encoding.}
We adopt a ResNet-50~\cite{he2016resnet}, which is pre-trained on ImageNet~\cite{deng2009imagenet}, with an FPN~\cite{lin2017feature} to obtain multi-scale image features. 
Following previous works~\cite{zhu2020deformabledetr,li2022bevformer}, the output features are from stage $S_{8\times}$, $S_{16\times}$ and $S_{32\times}$ of ResNet-50, where the subscripts $n\times$ indicates the downsampling factor.
In the FPN module, the features are transformed into a four-level output with an additional $S_{64\times}$ level. 
The number of output channels of each level is set to 256. 
Then we adopt a simplified view transformer with 3 encoder layers proposed in BEVFormer~\cite{li2022bevformer}. 
Note that we do not use temporal information, and thus the temporal self-attention layer in the BEVFormer encoder is replaced by a deformable attention~\cite{zhu2020deformabledetr} layer.
The size of BEV grids is set to $200 \times 100$, with four different height levels of $\{-1.5m, -0.5m, +0.5m, +1.5m\}$ relative to the ground.

\smallskip
\noindent
\textbf{Deformable Decoder.}
For the decoder, we utilize the decoder layer in Deformable DETR~\cite{zhu2020deformabledetr} that each decoder layer contains three layers: 
a self-attention layer with 8 attention heads, 
a deformable attention layer with 8 attention heads and 4 offset points,
and a two-layer feed-forward network with 512 channels in the middle. 
After each operation, a dropout layer with a ratio of 0.1 and a layer normalization is applied. 
The dimension of initial queries $q=[q_p, q_o] \in Q$ is set to 256, where $q_p$ is utilized to generate the initial reference point, and $q_o$ is the initial object query.
Notably, the reference points will remain unchanged across different layers.
The query number for centerlines and traffic elements is set to 200 and 100 respectively.

\smallskip
\noindent
\textbf{Scene Graph Neural Network.}
We utilize a simplified version of Graph Convolutional Network (GCN)~\cite{kipf2016semi} as our GNN layer. 
Given an input matrix $Q^{i} \in \mathbb{R}^{N \times C}$, with $N$ representing the number of nodes and $C$ denoting the number of channels, the output of the operation is:
\begin{equation}
    Q^{i'} = \sigma \Big( T^i Q^{i} \mathbf{W}^{i} \Big),
\end{equation}
where $\mathbf{W}^i \in \mathbb{R}^{C \times C}$ is the learnable weight matrix, $T^i \in \mathbb{R}^{N' \times N}$ describes the adjacency matrix with $N'$ output nodes, and $\sigma(\cdot)$ is the activation function. 
Note that the matrix $T$ is inferred without gradients during training.
For the traffic element branch, an embedding network is employed before each GNN layer. The embedding network is a two-layer MLP, in which the output channels are 512 and 256.
In between the MLP, a ReLU activation function and a dropout layer are included. $\beta_{ll}$ and $\beta_{lt}$ are set to 0.6.

\smallskip
\noindent
\textbf{Prediction Heads.}
The prediction head for perception comprises a classification head and a regression head.
For the traffic element branch, the classification head is a single-layer MLP, which outputs the sigmoid probability of each class. 
The regression head is a three-layer MLP with ReLU as the activation function in between, which predicts the normalized coordinates of 2D bounding boxes in the form of $\{cx, cy, width, height\}$.
For centerline, the classification head consists of a three-layer MLP with LayerNorm and ReLU in between, which predicts the confidence score. 
The regression head is a three-layer MLP with ReLU, which predicts the normalized point set of $11 \times 3$ for a centerline.
To predict topology relationships, relationship heads are applied.
Given the instance queries $\widetilde{Q}_{a}$ and $\widetilde{Q}_{b}$ with 256 feature channels, the topology head first applies a three-layer MLP:
\begin{equation}
    {\widetilde{Q}'_{a}} = \texttt{MLP}_{\texttt{a}} (\widetilde{Q}_{a}), \ {\widetilde{Q}'_{b}} = \texttt{MLP}_{\texttt{b}} (\widetilde{Q}_{b}),
\end{equation}
where the number of output channels is 128.
For each pair of queries $\tilde{q}'_{a} \in \widetilde{Q}'_a$ and $\tilde{q}'_{b} \in \widetilde{Q}'_b$, the output is the confidence of the relationship:
\begin{equation}
    conf. = \texttt{sigmoid} \Big( \texttt{MLP}_{\texttt{top}} \big( \texttt{concat}(\tilde{q}'_{a}, \tilde{q}'_{b}) \big) \Big),
\end{equation}
where the $\texttt{MLP}$ is independent in different types of relationships.

\smallskip
\noindent
\textbf{Loss.}
The $\mathcal{L}_{det_{\text{TE}}}$ is decomposed into a classification, a regression, and an IoU loss:
\begin{equation}
    \mathcal{L}_{det_{\text{TE}}} = \lambda_{cls} \cdot \mathcal{L}_{cls} + \lambda_{reg} \cdot \mathcal{L}_{reg} + \lambda_{iou} \cdot \mathcal{L}_{iou},
\end{equation}
where $\lambda_{cls}$, $\lambda_{reg}$, and $\lambda_{iou}$ is set to 1.0, 2.5, and 1.0 respectively. 
The classification loss $\mathcal{L}_{cls}$ is a Focal loss.
Note that the regression loss $\mathcal{L}_{reg}$ is an L1 Loss calculated on a normalized format of $\{cx, cy, width, height\}$, while the IoU loss $\mathcal{L}_{iou}$ is a GIoU loss computed on the denormalized coordinates.
For centerline detection, the $\mathcal{L}_{det_{\text{LC}}}$ comprises a classification and a regression loss:
\begin{equation}
    \mathcal{L}_{det_{\text{LC}}} = \lambda_{cls} \cdot \mathcal{L}_{cls} + \lambda_{reg} \cdot \mathcal{L}_{reg},
\end{equation}
where $\lambda_{cls}$ and $\lambda_{reg}$ is set to $1.5$ and $0.025$ respectively. 
Note that the regression loss is calculated on the denormalized 3D coordinates.
For topology reasoning, we adopt the same Focal loss but different weights on different types of relationships:
\begin{equation}
    \mathcal{L}_{top} = \lambda_{top_{ll}} \cdot \mathcal{L}_{top_{ll}} + \lambda_{top_{lt}} \cdot \mathcal{L}_{top_{lt}},
\end{equation}
where both $\lambda_{top_{ll}}$ and $\lambda_{top_{lt}}$ are set to $5.0$.

\smallskip
\noindent
\textbf{Training.}
The resolution of input images is 2048 $\times$ 1550, except for the front-view image, which is in the size of 1550 $\times$ 2048 and is cropped into 1550 $\times$ 1550.
For data augmentation, resizing with a factor of 0.5 and color jitter is used by default.
We adopt the AdamW optimizer~\cite{adamw} and a cosine annealing schedule with an initial learning rate of $1\,\times\,10^{-4}$. \algname is trained for 24 epochs with a batch size of 8 with 8 NVIDIA Tesla A100 GPUs.

\subsubsection{Re-implementation of SOTA Methods}

\begin{table}[t!]
    \centering
    \scalebox{0.85}{
        \begin{tabular}{c|l|ccccc}
    		\toprule
      
                Data & Method
                & DET$_{l}$$\uparrow$ 
                & TOP$_{ll}$$\uparrow$
                & DET$_{t}$$\uparrow$ 
                & TOP$_{lt}$$\uparrow$ 
                & OLS$\uparrow$ 
                \\
                
            \midrule
            
                \multirow{5}{*}{\rotatebox[origin=c]{90}{\parbox[t]{1.5cm}{\centering $subset\_A$}}}
                & STSU~\cite{can2021stsu}  & 12.7  & 0.5& 43.0 &  \underline{15.1} &  25.4\\
                & VectorMapNet~\cite{liu2022vectormapnet}  & 11.1  & 0.4& 41.7 &  5.9 & 20.8\\
                & MapTR~\cite{liao2023maptr} & 8.3&  0.2  & \underline{43.5} & 5.9 & 20.0 \\
                & MapTR*~\cite{liao2023maptr} &  \underline{17.7}&  \underline{1.1}&  \underline{43.5}&  10.4& \underline{26.0} \\
                & \textbf{\algname} (Ours) & \textbf{28.5} & \textbf{4.1} & \textbf{48.1} & \textbf{20.8} & \textbf{35.6}  \\
                
            \midrule
            
                \multirow{5}{*}{\rotatebox[origin=c]{90}{\parbox[t]{1.5cm}{\centering $subset\_B$}}}
                & STSU~\cite{can2021stsu}  & 8.2 &  0.0 & 43.9 & \underline{9.4} & 21.2\\
                & VectorMapNet~\cite{liu2022vectormapnet}  & 3.5  & 0.0& 49.1 &  1.4 & 16.3\\
                & MapTR~\cite{liao2023maptr}  & 8.3& 0.1 &  \underline{54.0}  & 3.7 & 21.1 \\
                & MapTR*~\cite{liao2023maptr} &  \underline{15.2}&  \underline{0.5}&  \underline{54.0}&  6.1& \underline{25.2} \\
                & \textbf{\algname} (Ours)  & \textbf{24.3} & \textbf{2.5} & \textbf{55.0} & \textbf{14.2} & \textbf{33.2}\\

            \bottomrule
    	\end{tabular}
    }
    \caption{
        \textbf{Comparison with state-of-the-art methods} on the \datasetname benchmark.
         \algname outperforms all previous works by a wide margin, especially in directed centerline perception and topology reasoning. 
        *: Topology reasoning evaluation is based on matching results on Chamfer distance.
        The highest score is bolded, while the second one is underlined.
    }
    \label{tab:ex:main}
    \vspace{-5px}
\end{table}

Since there are no prior methods for the task of driving scene understanding, we adapt three state-of-the-art algorithms which are initially designed for lane graph estimation or map learning: STSU~\cite{can2021stsu}, VectorMapNet~\cite{liu2022vectormapnet}, and MapTR~\cite{liao2023maptr}.
For a fair comparison, we utilize the same image backbone and adopt the same TE head as in \algname. %
As for topology reasoning, we treat it differently based on their own modeling concepts.
Specifically, STSU predicts centerlines as Bezier curves and their relationships. Thus most of its designs are retained, while other irrelevant modules are removed. Additionally, we append a topology head for predicting correspondence between centerlines and traffic elements.
The goal of VectorMapNet and MapTR is to predict vectorized map elements such as lanelines and pedestrian crossings. VectorMapNet utilizes a DETR-like decoder to estimate key points and then introduces an auto-regressive module to generate detailed graphical information for a laneline instance. MapTR directly predicts polylines with a fixed number of points using a Transformer decoder. 
To adapt them to our task, we supervise VectorMapNet and MapTR with centerline labels and use the element queries in their Transformer decoders as the instance queries to produce the driving scene topology.

\smallskip
\noindent
\textbf{STSU.}
The original model utilizes EfficientNet-B0~\cite{tan2019efficientnet} as the backbone and detects centerlines and vehicles under the monocular setting. 
It employs a BEV positional embedding and a DETR head to predict three Bezier control points for each centerline, and uses object queries in the decoder to predict the connectivity of centerlines.
To adapt to the multi-view inputs, we re-implement STSU by replacing the original backbone with ResNet-50 and aligning the image input resolution with our model. 
Subsequently, we compute and concatenate the BEV embedding of images from different views. 
The concatenated embedding is then fed into the DETR encoder. 
We retain the original DETR decoder to predict the Bezier control points, which are interpolated into 11 equidistant points as outputs. 
The lane-lane relationship prediction head of STSU is preserved as well. 
For the newly added traffic element detection branch and lane-traffic element topology head, we adopt the same design as in \algname to ensure a fair comparison.

\smallskip
\noindent
\textbf{VectorMapNet} utilizes a DETR-like decoder to estimate key points and an auto-regressive module to generate detailed graphical information for a laneline instance.
It uses a smaller input image size with a shorter edge length of 256.
We increase the length to 775 while maintaining the aspect ratio.
The backbone setting is aligned with \algname.
The perception range is defined as $\pm30\textit{m} \times \pm15\textit{m}$ in the original setting, but we expand it to
$\pm50\textit{m} \times \pm25\textit{m}$.
The centerline outputs of VectorMapNet are interpolated to fit our setting during the prediction process.
For topology prediction, we use the key point object queries in the VectorMapNet decoder as instance queries of centerlines. 
We implement the modification on the given codebase of VectorMapNet while retaining other settings.
However, due to their lack of support for 3D centerlines, we only predict 2D centerlines in the BEV space and ignore the height dimension.

\smallskip
\noindent
\textbf{MapTR.}
We align the original backbone setting with \algname. 
The perception region is expanded to $\pm50\textit{m} \times \pm25\textit{m}$.
We adopt MapTR to predict 11 points for each centerline. 
For topology prediction, we use the average of point queries of an instance in the MapTR decoder as the object query of a centerline. The traffic element head and the topology head are with the same setting as in \algname. 
The implementation is also done on the open-source codebase of MapTR.
Due to the lack of support for 3D centerlines, we only predict 2D centerlines in BEV and ignore the height dimension during inference.

\begin{table}[t!]
    \centering
    \scalebox{0.85}{
        \begin{tabular}{lc|cc|c|c}
    		\toprule

                Method  
                & Topology 
                & DET$_{l}$$\uparrow$ 
                & TOP$_{ll}$$\uparrow$
                & DET$_{l, chamfer}$$\uparrow$ 
                & FPS
                \\
                
            \midrule
            
                STSU~\cite{can2021stsu} & \ding{51} & 14.2 & 0.6 & 13.8 & \textbf{12.8} \\
                VectorMapNet~\cite{liu2022vectormapnet}  &  \ding{55} & 12.7 & - & 10.3 & 1.0 \\
                MapTR~\cite{liao2023maptr}  & \ding{55}  & 10.0 & - & 21.7 & 11.5 \\

            \midrule
            
                \textbf{\algname}  & \ding{51} & \textbf{27.7} & \textbf{4.6} & \textbf{27.4} & 10.1 \\

            \bottomrule
    	\end{tabular}
    }
    \caption{
        \textbf{Comparison on centerline perception with a ResNet-50 backbone.}
        ``Topology'' denotes that the network is trained with topology supervision.
    }
    \label{tab:ex:unifiedbackbone}
\end{table}

\begin{table}[t!]
    \centering
    \scalebox{0.85}{
    \begin{tabular}{l|c}
        \toprule
            Method & mIoU$\uparrow$ 
            \\
            \midrule
            HDMapNet~\cite{li2021hdmapnet} & 18.3 \\
            STSU~\cite{can2021stsu} & 31.1  \\
            VectorMapNet~\cite{liu2022vectormapnet} & 25.0  \\
            MapTR~\cite{liao2023maptr} & 35.7  \\
            \textbf{\algname} (Ours) & \textbf{39.0}  \\
            \bottomrule
    \end{tabular}
    }
    \caption{\textbf{Comparison on BEV segmentation.} When rendering centerlines on the BEV grids, \algname also outperforms the previous approach.}
    \label{tab:sota-seg}
    \vspace{-5px}
\end{table}

\subsubsection{Dataset and Metrics}

We conduct experiments on the \datasetname benchmark~\cite{openlanev2}, which covers complex urban scenarios. 
The dataset contains topological structures in the driving scenes, and raises huge challenges for algorithms to perceive and reason about the environment accurately.
Ablation studies are conducted on the $subset\_A$ of \datasetname.

\smallskip
\noindent
\textbf{Dataset.}
Built on top of the Argoverse 2~\citep{wilson2av2} and nuScenes~\citep{caesar2020nuscenes} datasets, the \datasetname benchmark includes images from 2,000 scenes collected worldwide under different environments.
The dataset is split into two subsets, namely $subset\_A$ and $subset\_B$.
Each subset contains 1,000 scenes with multi-view images and annotations at 2$Hz$.
All lanes within $[-50m, +50m]$ along the x-axis and $[-25m, +25m]$ along the y-axis are annotated in the 3D space.
Centerlines are described in the form of lists of points.
Each list is ordered and comprises 201 points in 3D space.
Statistically, about 90\% of frames have more than 10 centerlines while about 10\% have more than 40.
Traffic elements follow the typical labeling style in 2D detection that objects are labeled as 2D bounding boxes on the front-view images.
Each element is denoted as a 2D bounding box on the front view image, with its attribute.
There are 13 types of attributes, including \textit{unknown}, \textit{red}, \textit{green}, \textit{yellow}, \textit{go\_straight}, \textit{turn\_left}, \textit{turn\_right}, \textit{no\_left\_turn}, \textit{no\_right\_turn}, \textit{u\_turn}, \textit{no\_u\_turn}, \textit{slight\_left}, and \textit{slight\_right}.
The topology relationships are provided in the form of adjacency matrices based on the ordering of centerlines and traffic elements.
In the adjacency matrices, an entry $(i, j)$ is positive (\ie, 1) if and only if the elements at $i$ and $j$ are connected.
Statistically, most of the lanes have one predecessor and successor. 
The number of connections is up to 7, as a lane leads to many directions in a complex intersection.
For the correspondence between centerlines and traffic elements, the majority have less than 5 correspondence, while a tiny number of elements have up to 12.

\smallskip
\noindent
\textbf{Perception Metrics.}
The $\text{DET}$ score is the typical mean average precision (mAP) for measuring instance-level perception performance.
Based on the Fr\'echet distances~\cite{eiter1994frechet}, the $\text{DET}_{l}$ score is averaged over match thresholds of $\mathbb{T} = \{1.0, 2.0, 3.0\}$:
\begin{equation}
    \text{DET}_{l} = \frac{1}{|\mathbb{T}|} \sum_{t \in \mathbb{T}} AP_t.
\end{equation}
Note that the defined BEV range is relatively large compared to other lane detection datasets, so accurate perception of lanes in the distance is hard. 
As a result, thresholds $\mathbb{T}$ are relaxed based on the distance between the lane and the ego car.
The $\text{DET}_{t}$ uses IoU as the similarity measure and is averaged over different attributes $\mathbb{A}$ of traffic elements:
\begin{equation}
    \text{DET}_{t} = \frac{1}{|\mathbb{A}|} \sum_{a \in \mathbb{A}} AP_a.
\end{equation}

\smallskip
\noindent
\textbf{Reasoning Metrics.}
The $\text{TOP}$ score is an mAP metric adapted from the graph domain.
Specifically, given a ground truth graph $G = (V, E)$ and a predicted one $\hat{G} = (\hat{V}, \hat{E})$, it builds a projection on the vertices such that $V = \hat{V}' \subseteq \hat{V}$, where the Fr\'{e}chet and IoU distances are utilized for similarity measure among lane centerlines and traffic elements respectively.
Inside the predicted $ \hat{V}'$, two vertices are regarded as connected if the confidence of the edge is greater than $0.5$.
Then the TOP score is the averaged vertice mAP between $(V, E)$ and $(\hat{V}', \hat{E}')$ over all vertices:
\begin{equation}
    \text{TOP} = \frac{1}{|V|} \sum_{v \in V} \frac{\sum_{\hat{n}' \in \hat{N}'(v)} P(\hat{n}') \mathbbm{1}(\hat{n}' \in N(v))}{|N(v)|},
\end{equation}
where $N(v)$ denotes the ordered list of neighbors of vertex $v$ ranked by confidence and $P(v)$ is the precision of the $i$-th vertex $v$ in the ordered list.
The $\text{TOP}_{ll}$ is for topology among centerlines on graph $(V_{l}, E_{ll})$, and the $\text{TOP}_{lt}$ for topology between lane centerlines and traffic elements on graph $(V_{l} \cup V_{t}, E_{lt})$.

\begin{table}[t!]
    \centering
    \scalebox{0.85}{
        \begin{tabular}{c|l|ccccc}
    		\toprule
      
                Method
                & DET$_{l}$$\uparrow$ 
                & TOP$_{ll}$$\uparrow$
                & DET$_{t}$$\uparrow$ 
                & TOP$_{lt}$$\uparrow$ 
                & OLS$\uparrow$ 
                \\
                
            \midrule
            
                Baseline & 25.7 & 4.0 & 47.2 & 20.6 & 34.6 \\
                + SG & 27.7 & 3.7 & 48.0 & 20.1 & 35.0 \\
                + SKG & \textbf{28.5} & \textbf{4.1} & \textbf{48.1} & \textbf{20.8} & \textbf{35.6}  \\

            \bottomrule
    	\end{tabular}
    }
    \caption{
        \textbf{Ablation on the design of scene graph neural network}. 
        ``SG'' represents the vanilla scene graph, and ``SKG'' is the enhanced SGNN with the proposed scene knowledge graph.
    }
    \label{tab:ab:gnn}
    \vspace{-5px}
\end{table}

\begin{table}[t!]
    \centering
    \scalebox{0.85}{
    	\begin{tabular}{l|ccccc}
    		\toprule
             Method & DET$_{l}$$\uparrow$ 
                & TOP$_{ll}$$\uparrow$ & DET$_{t}$$\uparrow$ & TOP$_{lt}$$\uparrow$ 
                & OLS$\uparrow$ 
                \\
                \midrule
                 LL only & 27.9  & 3.8 & 47.8 & 20.3 & 35.1 \\
                 LT only & 27.8  & 3.9 & 47.5 & 20.5 & 35.1\\
                 \algname & \textbf{28.5} & \textbf{4.1} & \textbf{48.1} & \textbf{20.8} & \textbf{35.6}  \\
                \bottomrule
    	\end{tabular}
    }
    \caption{\textbf{Ablation on feature propagation} in the scene knowledge graph.
    ``LL only'' denotes that SGNN only aggregates the spatial information from lane-lane connectivity, and ``LT only'' indicates that only semantic information from the lane-traffic element relationship is included in SGNN.
    }
    \label{tab:ab:featurepropagation}
\end{table}

\begin{table}[t!]
    \centering
    \scalebox{0.85}{
    	\begin{tabular}{l|cccccc}
    		\toprule
                Method & DET$_{l}$$\uparrow$ 
                & TOP$_{ll}$$\uparrow$ & DET$_{t}$$\uparrow$ & TOP$_{lt}$$\uparrow$ 
                & OLS$\uparrow$ 
                \\
                \midrule
                w/o embedding & 28.4  & \textbf{4.1} & 46.9 & 20.5 & 35.2 \\
                \algname & \textbf{28.5} & \textbf{4.1} & \textbf{48.1} & \textbf{20.8} & \textbf{35.6}  \\
                \bottomrule
    	\end{tabular}
     }
    \caption{\textbf{Ablation on traffic element embedding.} TE embedding is necessary to deal with inconsistency in the feature space of different queries.}
    \label{tab:ablation-embedding}
    \vspace{-5px}
\end{table}

\smallskip
\noindent
\textbf{Overall Metrics.}
The primary task of the dataset is scene structure perception and reasoning, which requires the model to recognize the dynamic drivable states of lanes in the surrounding environment.
The OpenLane-V2 Score (OLS) summarizes metrics covering different aspects of the primary task:
\begin{equation}
    \text{OLS} = \frac{1}{4} \bigg[ \text{DET}_{l} + \text{DET}_{t} + f(\text{TOP}_{ll}) + f(\text{TOP}_{lt}) \bigg],
\end{equation}
where DET and TOP describe performance on perception and reasoning respectively, and $f$ is the square root function. 

\subsection{Main Results}
\label{sec:exp-results}

\begin{figure*}[t!]
  \centering
       \begin{subfigure}[b]{\linewidth}
         \centering
         \includegraphics[width=\linewidth]{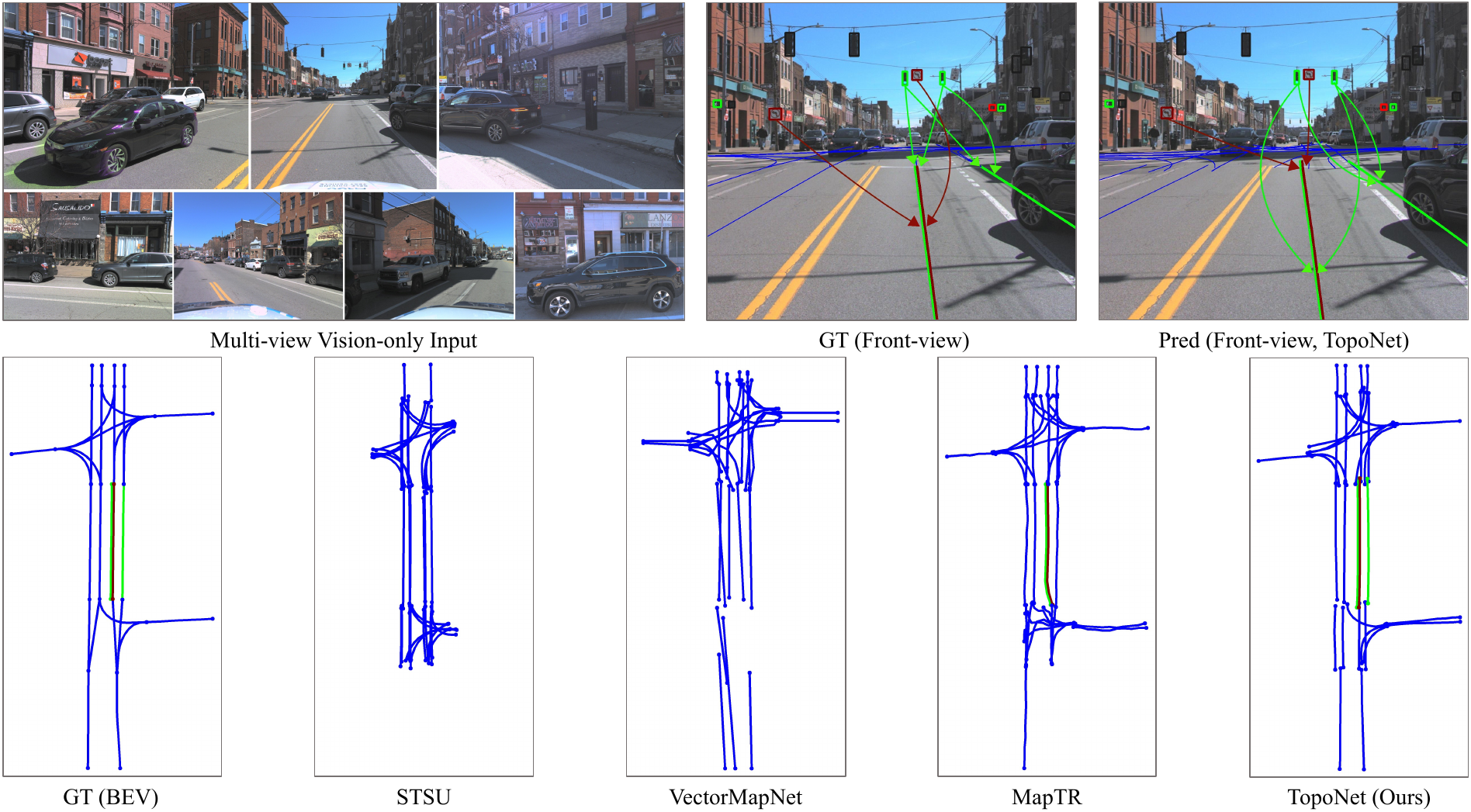}
         \vspace{-15px}
        \caption{Qualitative results on $subset\_A$ of the \datasetname dataset.}
        \vspace{3px}
     \end{subfigure}
     
          \begin{subfigure}[b]{\linewidth}
         \centering
         \includegraphics[width=\linewidth]{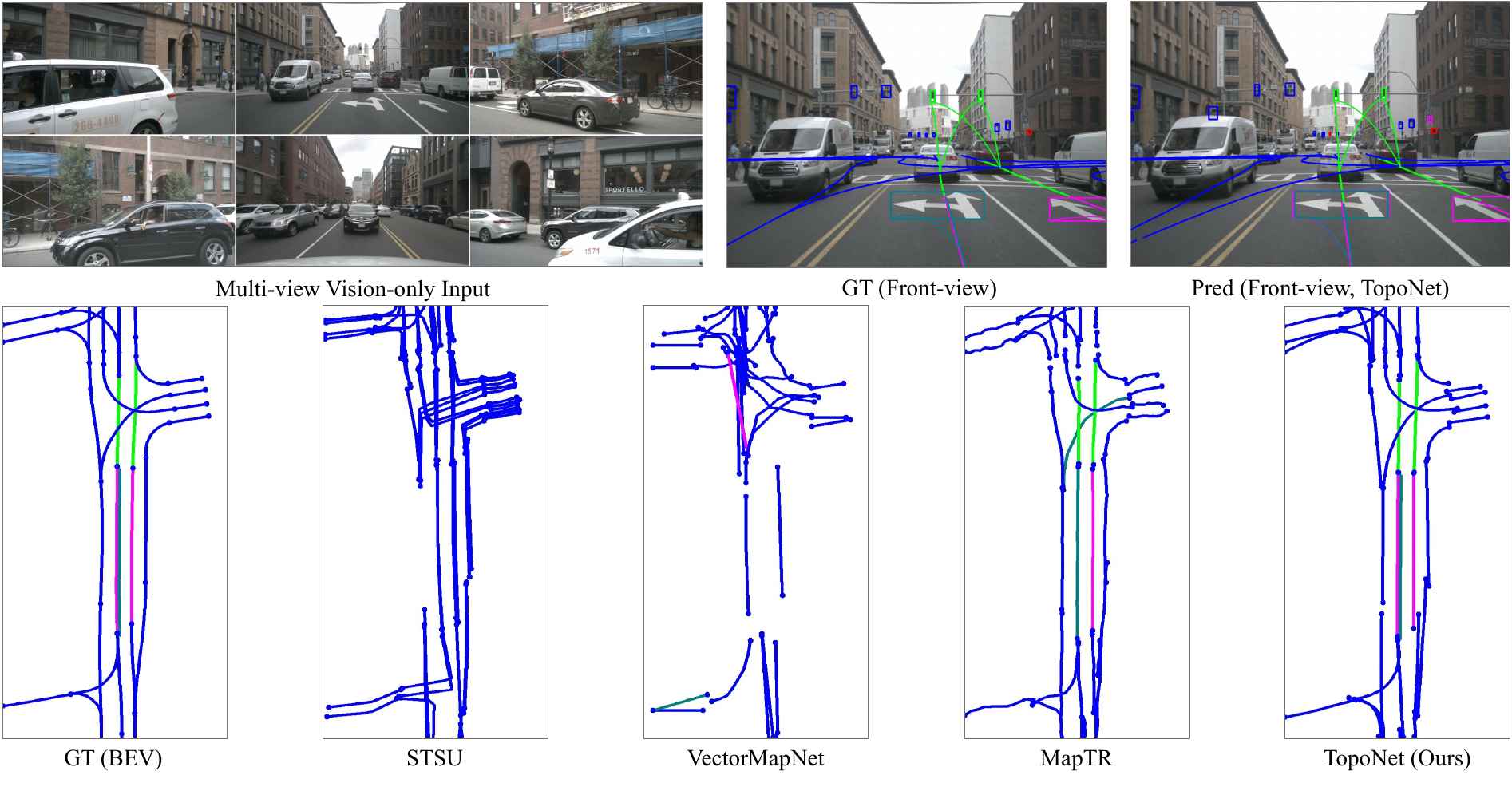}
         \vspace{-17px}
        \caption{Qualitative results on $subset\_B$ of the \datasetname dataset.}
     \end{subfigure}
      \vspace{-15pt}
  \caption{
  \textbf{Qualitative results} of \algname and other algorithms. While driving in complex scenarios, \algname achieves superior lane graph prediction performance compared to other SOTA methods. It also successfully builds all connections between traffic elements and lanes (top right, and correspondingly colored lines in BEV).
  Colors denote categories of traffic elements.
  }
  \label{fig:vis}
\end{figure*}

We compare the proposed \algname to several state-of-the-art methods in \Cref{tab:ex:main}. 
\algname outperforms all previous algorithms by a large margin.
As the SOTA map learning method MapTR ignores the direction of centerlines with the permutation-equivalent modeling, we additionally evaluate MapTR based on Chamfer distance matching. 
However, its performance on DET$_l$, as well as topology metrics, significantly degenerates. 
The performance of centerline queries without directional information indicates that understanding the complex scenario and perceiving presented instances are two totally different stories.
All methods achieve similar DET$_t$, since we adopt the same traffic element detection branch.
However, \algname still has a better performance on LC-TE topology reasoning, indicating the necessity of the proposed SGNN module, in which different entities are modeled differently.

\smallskip
\noindent
\textbf{Comparison on Centerline Perception.} 
To have a fair comparison, we use a unified backbone architecture and PV-to-BEV transformation module for various SOTA methods on centerline perception task.
We keep the topology supervision for STSU, as it is originally designed for detecting centerlines and their topology relationship.
Since VectorMapNet and MapTR are for the task of laneline detection where there is no relationship between visible lanelines, we alter the supervision from laneline to centerline and ignore topology supervision to preserve their design choice.
As shown in \Cref{tab:ex:unifiedbackbone}, \algname outperforms other methods, indicating the effectiveness of the model design.
To better align with previous works~\cite{liu2022vectormapnet, liao2023maptr}, we also provide DET$_{l,\text{chamfer}}$ with the Chamfer distance as the similarity measure.
It does not take the lane direction into account and is thresholded on \{0.5, 1.0, 1.5\}.
Under this metric, \algname also shows a prevailing performance compared to other methods.
Besides, we measure the runtime of each model on an A100 bare machine.
The FPS of \algname is 10.1. Compared to other methods on the same machine with aligned input size $512\!\times676$, our method has comparable online efficiency but higher performance.

\smallskip
\noindent
\textbf{Comparison on BEV Segmentation.} 
The prediction results of our method are rendered to BEV with a fixed line width of 0.75$m$ aligned with HDMapNet~\cite{li2021hdmapnet}, and the mIoU metric is adopted following common segmentation tasks. 
We reproduce HDMapNet with an enlarged BEV perception range of $\pm50\textit{m} \times \pm25\textit{m}$ to get aligned with \algname.
As shown in \Cref{tab:sota-seg}, \algname significantly surpasses other methods in terms of mIoU, verifying the advance of our framework.

\begin{figure*}[t!]
  \centering
    \includegraphics[width=\linewidth]{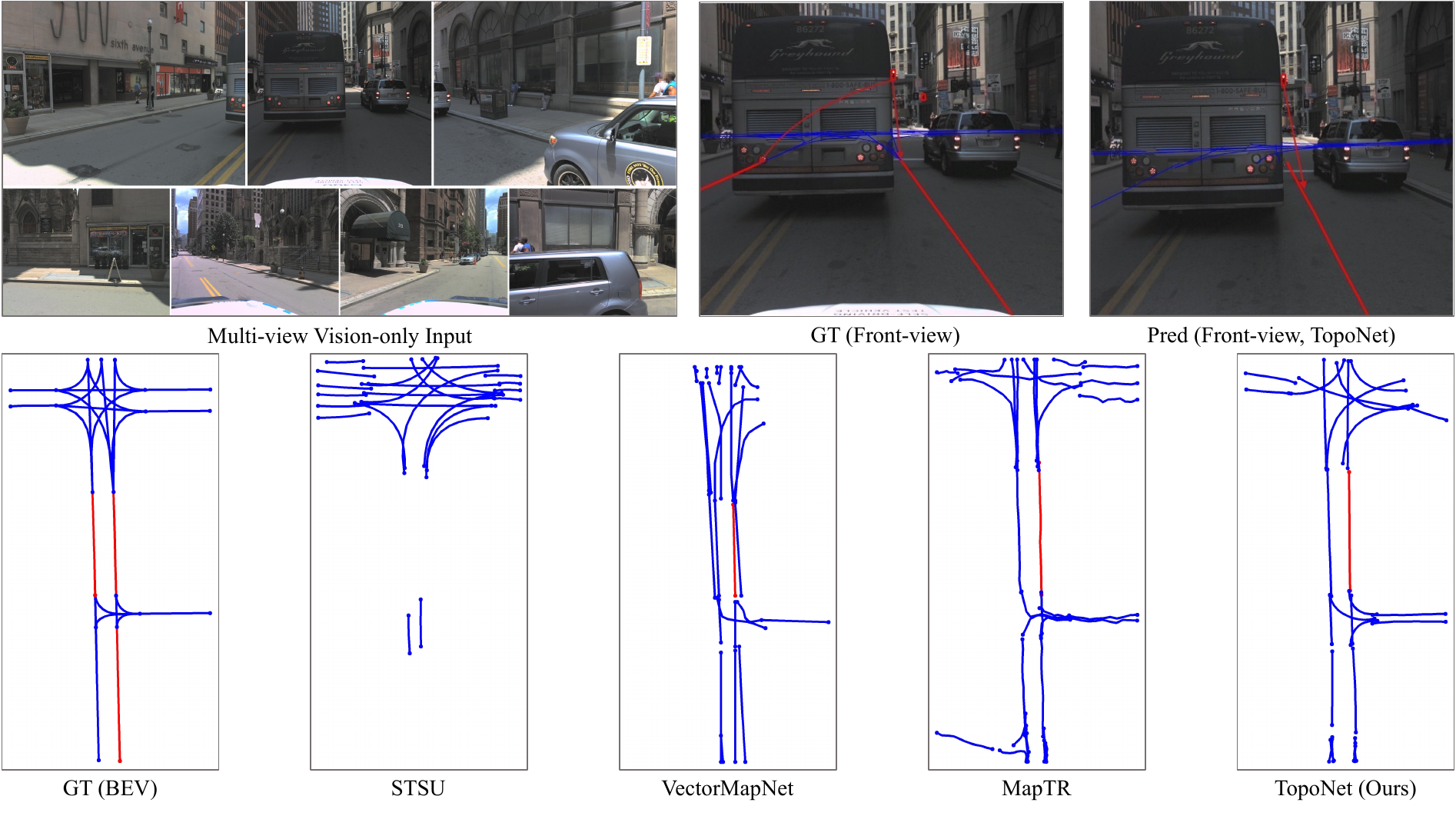}
     \vspace{-15pt}
  \caption{
  \textbf{Failure case under large-area occlusion.} \algname fails to predict centerlines and the lane graph in the intersection with a large bus colluding in front. Note that the relationship between the left lane and the red light is an incorrect annotation where our algorithm reasons about the direction of the left lane and avoids the false positive prediction.
  }
  \label{fig:supp-vis-occlusion}
\end{figure*}

\begin{table}[t!]
    \centering
    \scalebox{0.85}{
    	\begin{tabular}{c|cccccc}
    		\toprule
                \# GNN & DET$_{l}$$\uparrow$ 
                & TOP$_{ll}$$\uparrow$ & DET$_{t}$$\uparrow$ & TOP$_{lt}$$\uparrow$ 
                & OLS$\uparrow$ 
                \\
                \midrule
                1 & \textbf{28.5} & \textbf{4.1} & \textbf{48.1} & \textbf{20.8} & \textbf{35.6}  \\
                2 & 27.9 & 4.0 & 47.5 & 20.9 & 35.3 \\
                3 & 20.4 & 0.5 & 46.1 & 15.7 & 28.3 \\
                4 & 12.2 & 0.0 & 48.5 & 8.2 & 22.6 \\
                \bottomrule
    	\end{tabular}
    
    }
    \caption{\textbf{Ablation on the number of GNN layers} in the scene knowledge graph. Model performance drops as the number of SGNN layers increases.}
    \label{tab:ab:gnnlayer}
    \vspace{-5px}
\end{table}

\subsection{Ablation Study}
\label{sec:exp-ablation}

\noindent
\textbf{Effect of Design in Scene Graph Neural Network.}
For ablations on the proposed SGNN, we alternate the proposed network into a baseline without feature propagation by downgrading the SGNN module to an MLP and the supervision of topology reasoning only occurs at the final decoder layer. The concatenation and down-sampling operations, as well as the traffic element embedding, are also removed. As illustrated in \Cref{tab:ab:gnn}, the proposed SKG outperforms models in other settings, demonstrating its effectiveness for topology understanding.
Compared to the SG version, the scene knowledge graph provides an additional improvement of 0.8\% for centerline perception, owning to the predefined semantic prior encoded in the categories of traffic elements.
The improvement of traffic element detection and topology reasoning is also consistent.

\smallskip
\noindent
\textbf{Effect on Feature Propagation.}
In the ``LL only'' setting, we set the $\beta_{lt}$ parameter to 0. Similar to the baseline, we remove the concatenation and down-sampling operations, as well as the traffic element embedding.
For ``LT only'', we set the $\beta_{ll}$ parameter to 0, while other modules remain intact. 
Results are reported in \Cref{tab:ab:featurepropagation}.
In the ``LL only'' setting, the drop on $\text{TOP}_{lt}$ demonstrates the importance of the graph $G_{lt}$.
With the ``LT only'' design, $\text{DET}_l$ degenerates when removing the graph $G_{ll}$, showing the importance of feature propagation between centerline queries.
These experiments show that both branches are necessary for achieving satisfactory model performance on the primary task.

\begin{table}[t!]
    \centering
    \scalebox{0.85}{
    	\begin{tabular}{c|cccccc}
    		\toprule
                Edge Weight & DET$_{l}$$\uparrow$ 
                & TOP$_{ll}$$\uparrow$ & DET$_{t}$$\uparrow$ & TOP$_{lt}$$\uparrow$ 
                & OLS$\uparrow$ 
                \\
                \midrule
                0.5 & 28.4  & 4.0 & 47.7 & \textbf{20.8} & 35.4 \\
                0.6 & \textbf{28.5} & \textbf{4.1} & \textbf{48.1} & \textbf{20.8} & \textbf{35.6}  \\
                0.7 & 27.3 & \textbf{4.1} & 47.7 & 20.7 & 35.1\\
                \bottomrule
    	\end{tabular}
    }
    \caption{\textbf{Ablation on edge weight} in the scene knowledge graph.}
    \label{tab:ab:weight}
    \vspace{-5px}
\end{table} 

\smallskip
\noindent
\textbf{Effect on Traffic Element Embedding.}
In the ``w/o embedding'' setting, we remove the traffic element embedding network and use $Q_t$ as the input of SGNN directly.
As shown in \Cref{tab:ablation-embedding}, removing the embedding results in a 1.2\% performance drop in traffic element recognition.
The reason is that TE queries contain a large amount of spatial information in the PV space due to the 2D detection supervision signals, resulting in significant inconsistencies in the feature spaces.
In all, the experiments demonstrate that TE embedding effectively filters out irrelevant spatial information and extracts high-level semantic knowledge to help centerline detection and lane topology reasoning.

\smallskip
\noindent
\textbf{Effect on the Number of GNN Layers.}
Though GNN is beneficial for propagating features in the knowledge graph, raising the number of GNN layers leads to degenerated performance. As shown in \Cref{tab:ab:gnnlayer}, SGNN with a single GNN layer achieves the best performance. The reason is that a GNN layer increases the similarity of adjacent vertices. When stacking multiple GNN layers, features of all vertices become less discriminative.

\smallskip
\noindent
\textbf{Effect on Edge Weight.}
Edge weight in the scene knowledge graph represents how much information is propagated through the SGNN layers. In \Cref{tab:ab:weight}, 0.6 corresponds to the most appropriate ratio and is used as the default value.

\subsection{Qualitative Analysis}

We provide a qualitative comparison in \cref{fig:vis}. \algname predicts most centerlines correctly and constructs a lane graph in BEV. Yet, prior works fail to output all entities or get confused about their connectivity. 
\cref{fig:supp-vis-occlusion} shows a case where a bus occludes the intersection in the front view image. 
\algname fails to predict lanes and the topology, especially those in the left half of the crossing. 
A large-scale dataset and learning techniques, such as active learning, would solve such failure cases in a real-world deployment.

\section{Conclusion and Future Work}

In this paper, we discuss abstracting driving scenes as topology relationships and propose the first resolution, namely \algname, to address the problem. Importantly, our method models feature interactions via the graph neural network architecture and incorporate traffic knowledge in heterogeneous feature spaces with the knowledge graph-based design.
Our experiments on the large-scale \datasetname benchmark demonstrate that \algname excels prior SOTA approaches on perceiving and reasoning about the driving scene topology under complex urban scenarios.

\smallskip
\noindent
\textbf{Limitations and Future Work.} Due to the query-based design for feature interactions, \algname performs well in achieving most positive predictions, while post-processes such as merging or pruning are still needed to produce clean output as in lane topology works~\cite{can2021stsu, buchner2023lanegnn}.
How to incorporate the merging ability with auto-regressive or other association mechanisms deserves future exploration. Meanwhile, it will be interesting to see if more categories of traffic elements, and correspondingly more sophisticated knowledge graphs will make any advances.

\section*{License of Assets}

The training data are from the \datasetname dataset~\cite{openlanev2}, which is under the CC-BY-NC-SA 4.0 license. The evaluation code in the \datasetname dataset is under the Apache-2.0 license. 
We integrate the pre-trained ResNet model and part of the BEVFormer~\cite{li2022bevformer} in our code, which are under the MIT license and Apache-2.0 license respectively.

\bibliographystyle{ieeetr}
\bibliography{bibliography_short, bibliography, bibliography_custom, bibliography_challenge}

\end{document}